\pdfoutput=1
\documentclass[11pt]{article}

\usepackage[]{acl}

\usepackage{times}
\usepackage{latexsym}
\usepackage{tabularx}
\usepackage{soul}

\usepackage[T1]{fontenc}
\usepackage[utf8]{inputenc}

\usepackage{microtype}
\usepackage{booktabs}

\usepackage{graphicx}
\usepackage{enumitem}

\title{Practical Benefits of Feature Feedback
Under Distribution Shift}

\author{%
  Anurag Katakkar*, Clay H. Yoo*, Weiqin Wang, \\ \textbf{Zachary C. Lipton, Divyansh Kaushik}\\
  Carnegie Mellon University\\
  \texttt{akatakka,hyungony,weiqinw,zlipton,dkaushik@cmu.edu}
}

\begin{document}
\maketitle
\begin{abstract}
In attempts to develop sample-efficient
and interpretable algorithms,
researcher have explored myriad mechanisms
for collecting and exploiting \emph{feature feedback} (or \emph{rationales})
auxiliary annotations provided
for training (but not test) instances
that highlight salient evidence.
Examples include bounding boxes around objects
and salient spans in text.
Despite its intuitive appeal, 
feature feedback has not delivered 
significant gains in practical problems
as assessed on iid holdout sets.
However, recent works on 
counterfactually augmented data
suggest an alternative benefit
of supplemental annotations, beyond 
interpretability:
lessening sensitivity to spurious patterns
and consequently delivering gains
in out-of-domain evaluations.
We speculate that while
existing methods 
for incorporating feature feedback
have delivered negligible in-sample 
performance gains,
they may nevertheless 
provide out-of-domain benefits.
Our experiments addressing sentiment analysis, 
show that feature feedback methods
perform significantly better on
various natural out-of-domain datasets
despite comparable in-domain evaluations.
By contrast, performance on natural language inference remains comparable. 
Finally, we compare those tasks
where feature feedback does (and does not) help.
% Add sentence somewhere about why/how this is relevant to blackbox nlp kinda topics
\end{abstract}

\section{Introduction}
Addressing various classification tasks
in natural language processing (NLP),
including sentiment analysis \citep{zaidan2007using}, 
natural language inference (NLI) \citep{deyoung2020eraser},
and propaganda detection \citep{pruthi2020weakly},
researchers have introduced resources 
containing additional side information
by tasking humans with marking spans in the input text 
(called \emph{rationales} or \emph{feature feedback)} 
that provide supporting evidence for the label.
For example, spans like ``underwhelming'', 
``horrible'', or ``worst film since Johnny English''
might indicate negative sentiment in a movie review.
Conversely, spans like ``exciting'', ``amazing'',
or ``I never thought Vin Diesel would make me cry''
might indicate positive sentiment.

These works have proposed a variety of strategies
for incorporating feature feedback
as additional supervision 
\citep{lei2016rationalizing,zhang2016rationale,lehman2019inferring, chen2019seeing,jain2020learning,deyoung2020eraser,pruthi2020weakly}.
Other researchers have studied 
the learning-theoretic properties
of feature feedback
\citep{poulis2017learning, dasgupta2018learning, dasgupta2020robust}.
We focus our study on the resources 
and practical methods developed for NLP.

Some have used this feedback 
to perturb instances for 
data augmentation \citep{zaidan2007using},
while others have explored multitask objectives 
for simultaneously classifying documents
and extracting rationales
\citep{pruthi2020weakly}.
A number of papers exploit feature feedback
as intermediate supervision 
for building extract-then-classify pipelines
\citep{chen2019seeing, lehman2019inferring, jain2020learning}.
One common assumption is that resulting models 
would learn to identify and rely more 
on spans relevant to the target labels, 
which would in turn lead to more accurate predictions.

However, despite their intuitive appeal, 
feature feedback methods have thus far
yielded underwhelming results on 
independent drawn and identically distributed (iid) test sets 
in applications involving deep nets.
While \citet{zaidan2007using} found
significant gains when incorporating rationales
into their SVM learning scheme,
benefits have been negligible in the BERT era.
For example, although \citet{pruthi2020weakly}
and \citet{jain2020learning}
address a different aim towards boosting interpretability---to 
improve extraction accuracy---their 
experiments show no improvement
in classification accuracy
by 
% virtue of 
incorporating rationales.

On the other hand, 
\citet{kaushik2020learning},
introduced counterfactually augmented data (CAD)
with the primary aim of showing 
how supplementary annotations
can be incorporated to make models
less sensitive to spurious patterns,
and additionally demonstrated
that models trained on CAD degraded less 
in a collection of out-of-domain tests
than their vanilla counterparts.
In followup work, they showed 
that for both CAD and feature feedback,
although corruptions to evidence spans via random word flips
result in performance degradation 
both in- and out-of-domain,
when non-evidence spans are corrupted,
out-of-domain performance 
often improves \citep{kaushik2021explaining}.
These findings echo earlier results in computer vision
\citep{ross2017right, ross2018improving}
where regularizing input gradients
(so-called \emph{local explanations})
to accord with expert attributions
led to an improved out-of-domain performance.

In this paper, we conduct an empirical study
of the out-of-domain benefits
of incorporating feature feedback in selected domains in NLP (sentiment analysis and NLI).
We seek to address two primary research questions:
(i) do models that rely on feature feedback 
generalize better out of domain 
compared to \emph{classify-only} models 
(i.e., models trained without feature feedback)? 
and (ii) do we need to solicit feature feedback 
for an entire dataset 
or can significant benefits be realized 
with a modest fraction of examples annotated?
Our experiments on sentiment analysis 
\citep{zaidan2007using}
and NLI \citep{deyoung2020eraser}
use both linear, BERT \citep{devlin2019bert}, and ELECTRA \citep{clark2019electra} models,
using two feature feedback techniques
\citep{pruthi2020weakly,jain2020learning}.

We limit our experiments to sentiment analysis 
and NLI only, although other tasks such as hate 
speech and propaganda detection
might appear to be natural candidates to include 
in our study as well. Hate Speech detection is 
an inherently subjective task. For 
example, \citep{Waseem2016AreYA} 
documented the disagreement between labels
collected from the crowd and those 
annotated by experts. Similarly, 
\citep{Ross2017MeasuringTR} documented 
that annotating hate speech itself is a 
hard task leading to low inter-rater 
agreement within the crowd as well. Thus, 
even though several hate speech 
classification datasets exist, in our 
view, they are not suitable for the 
research questions we ask in the 
paper---what might be labeled as hate by 
one annotator may not be labeled hate in 
another dataset by another annotator, 
making it difficult to attribute
the impact on performance
to generalization ability or
some other factors (such as noisy labeling, 
or choice of labeling instructions, etc.). 
As for propaganda detection, while a 
dataset with high-quality labels 
and feature 
feedback annotations exists, the lack of 
additional datasets 
restricts our ability to train and evaluate 
the resulting models on a battery of 
out-of-domain datasets.

We find that sentiment analysis models 
fine-tuned with feature feedback on IMDb data
see no improvement in in-domain accuracy.
However, out-of-domain, 
sentiment analysis models benefit 
significantly from feature feedback.
For example, ELECTRA and BERT models
both see gains of $\approx 6\%$ 
on both Amazon \citep{ni2019justifying} 
and Yelp reviews \citep{kaushik2021explaining}
even when feature feedback is available for just $25\%$ of instances. 
However, on NLI, we find that
both iid and out-of-domain performance
are comparable with or without feature feedback.
We further find that while 
for sentiment analysis,
rationales constitute only $\approx21\%$ 
of all unique tokens in the training set,
for NLI they constitute $\approx80\%$,
potentially helping to explain 
why feature feedback is less useful there.

\section{Methods and Datasets}

% Our study centers 
We focus on two techniques
(classify-and-extract~\citep{pruthi2020weakly} 
and extract-then-classify~\citep{jain2020learning}),
two pretrained models,
and one (in-domain) dataset each
for sentiment analysis and NLI
that contain feature feedback.
% \paragraph{Techniques for Incorporating Feature Feedback} 
% 
% classify-and-extract model~\citep{pruthi2020weakly} 
% and extract-then-classify model~\citep{jain2020learning}.
% Both these models have roughly achieved
% state-of-the-art classification performance,
% with \citep{pruthi2020weakly} achieving
% significantly better extraction performance.
For both techniques,
\emph{feature feedback} annotations
provide supervision to the extractive component.
The classify-and-extract model
jointly predicts the (categorical) label % categorical target variable
and performs sequence tagging predict rationales.
The classification head and a linear chain CRF~\citep{lafferty2001conditional}
share an encoder, initialized with pretrained weights. 

The extract-then-classify method \citep{jain2020learning}
first trains a classifier (\emph{support}) on complete examples 
to predict the label, using its
outputs to extract continuous feature importance scores.
These scores are then binarized
using a second classifier (\emph{extractor})
which is trained on the feature importance scores
from \emph{support} and makes
token-level binary predictions
to identify rationale tokens in the input.
A binary cross-entropy term 
in the objective of the extractor 
is used to maximise agreement
of the extracted tokens with human rationales.
Finally, a third classifier (\emph{predictor}) is trained
to predict the target (sentiment or entailment) label
based only on these extracted tokens.

For both approaches, we experiment 
with two pretrained models (BERT and ELECTRA).
We limit the maximum sequence length to $512$ tokens
and train all models for $10$ epochs 
using AdamW optimizer \citep{loshchilov2019decoupled}
with a learning rate of $2e-5$ 
and a batch size of $8$ 
and early stopping based on mean 
of classification and extraction F$1$ scores 
on the validation set.
We replicate all experiments on $5$ seeds
and report mean performance
along with standard deviation.

\begin{table}[t!]
\centering
    \small
\begin{tabular}{l c c c}
\toprule
Test set &  Classify-only & \citeauthor{pruthi2020weakly}  & \citeauthor{jain2020learning}\\
\midrule
\multicolumn{4}{c}{BERT}\\
\midrule
In-domain & $85.9_{0.7}$ & $\mathbf{89.9_{2.3}}$ & $\mathbf{90.4_{0.3}}$\\
CRD & $89.3_{0.7}$ & $\mathbf{91.6_{0.7}}$ & $87.5_{0.8}$\\
SST2 & $77.6_{4.1}$ & $79.3_{3.6}$ & $75.6_{1.2}$\\
Amazon & $78.1_{4.9}$ & $83.5_{3.1}$ & $\mathbf{92.3_{1.2}}$\\
Semeval & $70.6_{5.7}$ & $73.2_{2.6}$ & $68.6_{2.2}$\\
Yelp & $86.8_{1.7}$ & $85.7_{1.6}$ & $\mathbf{91.6_{0.1}}$\\ 
\midrule
\multicolumn{4}{c}{ELECTRA}\\
\midrule
In-domain & $93.2_{0.3}$ & $91.8_{1.4}$ & $93.1_{0.3}$\\
CRD & $91.6_{0.4}$ & $\mathbf{93.7_{0.9}}$ & $91.5_{0.7}$\\
SST2 & $73.2_{1.3}$ & $74.0_{1.2}$ & $\mathbf{77.2_{1.4}}$\\
Amazon & $72.8_{2.0}$ & $75.5_{2.1}$ & $\mathbf{84.2_{1.6}}$\\
Semeval & $67.5_{4.5}$ & $72.5_{1.8}$ & $66.7_{3.0}$\\
Yelp & $79.0_{3.6}$ & $\mathbf{84.6_{1.8}}$ & $\mathbf{94.7_{0.2}}$\\
\bottomrule
\end{tabular}
\caption{Mean and standard deviation (in subscript) of accuracy scores of classify-only models, and models proposed by \citet{pruthi2020weakly} and \citet{jain2020learning}, fined-tuned for sentiment analysis. Significant results ($p<0.05$) compared to the classify-only models are highlighted in bold. \label{tab:sentiment1}}
\end{table}

% To verify that our hypothesis holds across
% To test our hypothesis across
To see whether results are consistent across architectures,
we also use a linear SVM \citep{zaidan2007using} 
with a modified objective function
on top of the ordinary soft-margin SVM, i.e.,
$$\frac{1}{2}||w||^2 + C(\sum_i \delta_i) + C\textsubscript{contrast}(\sum_{i,j} \xi_{ij})$$
subject to the constraints
$\vec{w} \cdot \vec{x}_{ij} \cdot y_i \geq 1 - \xi_{ij} \; \forall i, j$
where 
$\vec{x}_{ij} := \frac{\vec{x}_i - \vec{v}_{ij}}{\mu}$
are \textit{psuedoexamples}, created 
by subtracting \textit{contrast-examples} ($\vec{v}_{ij}$),
input sentence void of randomly chosen rationales,
from the original input ($\vec{x}_i$).
We use term-frequency embeddings with unigrams
appearing in at least $10$ reviews
and set $C = C_{contrast} = \mu = 1$. 
For each training example,
we generate $5$ psuedoexamples.

\begin{table}[t!]
\centering
    \small
\begin{tabular}{l c c c}
\toprule
Test set &  Classify-only & \citeauthor{pruthi2020weakly}  & \citeauthor{jain2020learning}\\
\midrule
\multicolumn{4}{c}{BERT}\\
\midrule
In-domain & $88.7_{2.0}$ & $89.8_{0.8}$ & $77.7_{0.1}$\\
RP & $62.9_{3.9}$ & $66.6_{0.6}$ & $57.9_{0.1}$\\
RH & $76.9_{3.5}$ & $80.5_{1.9}$ & $70.7_{0.2}$\\
MNLI-M & $69.7_{2.6}$ & $68.1_{1.9}$ & $69.8_{0.1}$\\
MNLI-MM & $71.5_{2.7}$ & $69.2_{2.3}$ & $66.2_{0.1}$\\
\midrule
\multicolumn{4}{c}{ELECTRA}\\
\midrule
In-domain & $\mathbf{96.0_{0.2}}$ & $95.0_{0.3}$ & $85.4_{0.04}$\\
RP & $80.8_{1.0}$ & $78.0_{0.6}$ & $72.2_{0.1}$\\
RH & $88.9_{1.0}$ & $88.7_{0.9}$ & $79.7_{0.1}$\\
MNLI-M & $86.5_{0.9}$ & $81.9_{2.1}$ & $77.1_{0.1}$\\
MNLI-MM & $86.6_{0.8}$ & $82.1_{2.0}$ & $75.7_{0.1}$\\
\bottomrule
\end{tabular}
\caption{Mean and standard deviation (in subscript) of F1 scores of models fine-tuned for NLI with an increasing number of examples with feature feedback. Significant results ($p<0.05$) compared to the classify-only models are highlighted in bold. \label{tab:nli2}}
% \vspace{-5mm}
\end{table}

%Anurag to again (probably?) talk about appropriateness of datasets
\paragraph{Datasets}
For sentiment analysis,
we use an IMDb movie reviews dataset \citep{zaidan2007using}.
Reviews in this dataset are labeled 
as having either \emph{positive} 
or \emph{negative} sentiment.
\citet{zaidan2007using} also tasked annotators 
to mark spans in each review 
that were indicative of the overall sentiment.
We use these spans as feature feedback.
Overall, the dataset has $1800$ reviews
in the training set (with feature feedback) 
and $200$ in test (without feature feedback). 
Since the test set does not include 
ground truth labels for evidence extraction, 
we construct a test set out of the $1800$ examples 
in the original training set. 
This leaves $1200$ reviews for a new training set, 
$300$ for validation, and $300$ for test.
For NLI, we use a subsample 
of the E-SNLI dataset \citep{deyoung2020eraser} 
used in \citet{kaushik2021explaining}. 
In this dataset, there are $6318$ premise-hypothesis pairs,
equally divided across \emph{entailment} and
\emph{contradiction} categories. 
% Each pair is accompanied with marked tokens 
% deemed relevant to the label's applicability.

We evaluate on CRD \citep{kaushik2020learning},
SST-2 \citep{socher2013recursive},
Amazon reviews \citep{ni2019justifying}, 
Tweets \citep{rosenthal2017semeval} 
and Yelp reviews \citep{kaushik2021explaining} 
for sentiment analysis, 
and Revised Premise (RP), 
Revised Hypothesis (RH) \citep{kaushik2020learning}, 
MNLI matched (MNLI-M) 
and mismatched (MNLI-MM) \citep{williams2018mnli} for NLI.

\begin{table*}[t!]
\centering
% \small
\begin{tabular}{l c c c c c}
\toprule
& \multicolumn{4}{c}{Fraction of Training Data with Rationales}\\
Evaluation set &  No rationales & $25\%$ & $50\%$ & $75\%$ & $100\%$ \\
\midrule
\multicolumn{6}{c}{BERT}\\
\midrule
In-domain & $85.9_{0.7}$ & $\mathbf{87.7_{1.1}}$ & $88.1_{2.4}$ & $\mathbf{90.2_{1.5}}$ & $\mathbf{89.9_{2.3}}$ \\
CRD & $89.3_{0.7}$ & $\mathbf{91.7_{0.6}}$ & $\mathbf{92.3_{0.9}}$ & $\mathbf{92.3_{0.3}}$ & $\mathbf{91.6_{0.7}}$ \\
SST2 & $77.6_{4.1}$ & $81.2_{0.6}$ & $81.3_{0.7}$ & $81.8_{0.6}$ & $79.3_{3.6}$ \\
Amazon & $78.1_{4.9}$ & $\mathbf{85.3_{1.2}}$ & $\mathbf{84.6_{1.7}}$ & $\mathbf{84.0_{0.5}}$ & $83.5_{3.1}$ \\
Semeval & $70.6_{5.7}$ & $\mathbf{77.8_{1.0}}$ & $75.5_{0.8}$ & $74.9_{0.8}$ & $73.2_{2.6}$\\
Yelp & $86.8_{1.7}$ & $86.9_{1.1}$ & $85.8_{1.5}$ & $85.4_{0.7}$ & $85.7_{1.6}$\\ 
\midrule
\multicolumn{6}{c}{ELECTRA}\\
\midrule
In-domain & $93.2_{0.3}$ & $92.4_{0.9}$ & $92.8_{1.2}$ & $93.7_{1.9}$ & $91.8_{1.4}$\\
CRD & $91.6_{0.4}$ & $92.1_{0.8}$ & $\mathbf{93.0_{0.6}}$ & $\mathbf{93.1_{0.3}}$ & $\mathbf{93.7_{0.9}}$\\
SST2 & $73.2_{1.3}$ & $73.1_{1.8}$ & $72.3_{1.6}$ & $72.3_{1.1}$ & $74.0_{1.2}$\\
Amazon & $72.8_{2.0}$ & $\mathbf{79.0_{1.8}}$ & $\mathbf{75.7_{1.2}}$ & $\mathbf{76.6_{1.8}}$ & $75.5_{2.1}$\\
Semeval & $67.5_{4.5}$ & $70.5_{1.5}$ & $66.2_{1.5}$ & $67.1_{2.2}$ & $72.5_{1.8}$\\
Yelp & $79.0_{3.6}$ & $\mathbf{84.5_{1.1}}$ & $\mathbf{84.2_{1.7}}$ & $\mathbf{84.3_{1.2}}$ & $\mathbf{84.6_{1.8}}$\\
\bottomrule
\end{tabular}
\caption{Mean and standard deviation (in subscript) of accuracy scores of models fine-tuned for sentiment analysis using the method proposed by \citet{pruthi2020weakly} with different base models (BERT and ELECTRA) and increasing proportion of examples with feature feedback. Results highlighted in bold are significant difference with $p<0.05$. \label{tab:sentiment2}}
\end{table*}

\begin{table*}[t!]
\centering
% \small
\begin{tabular}{l c c c c c}
\toprule
& \multicolumn{4}{c}{Fraction of Training Data with Rationales}\\
Evaluation set &  No rationales & $25\%$ & $50\%$ & $75\%$ & $100\%$ \\
\midrule
\multicolumn{6}{c}{BERT}\\
\midrule
In-domain & $88.7_{2.0}$ & $89.6_{0.4}$  & $89.9_{0.4}$ & $89.7_{0.4}$ & $89.8_{0.8}$ \\
RP & $62.9_{3.9}$ & $67.6_{2.0}$  & $67.4_{1.2}$ & $68.6_{0.6}$ & $66.6_{0.6}$ \\
RH & $76.9_{3.5}$  & $80.4_{1.1}$ & $81.7_{1.6}$ & $81.4_{0.7}$ & $80.5_{1.9}$ \\ 
MNLI-M & $69.7_{2.6}$ & $67.6_{3.4}$ & $68.1_{4.6}$ & $68.8_{2.0}$ & $68.1_{1.9}$ \\ 
MNLI-MM & $71.5_{2.7}$  & $68.8_{4.5}$ & $69.2_{5.9}$ & $69.8_{2.7}$ & $69.2_{2.3}$ \\ 
\midrule
\multicolumn{6}{c}{ELECTRA}\\
\midrule
In-domain & $\mathbf{96.0_{0.2}}$ & $95.1_{0.3}$  & $95.0_{0.3}$ & $95.0_{0.3}$ & $95.0_{0.3}$ \\
RP & $80.8_{1.0}$ & $78.2_{1.3}$  & $79.2_{1.1}$ & $77.2_{1.3}$ & $78.0_{0.6}$ \\
RH & $88.9_{1.0}$ & $88.0_{1.2}$ & $88.4_{0.3}$ &  $87.9_{0.4}$ & $88.7_{0.9}$ \\ 
MNLI-M & $86.5_{0.9}$ & $82.0_{2.8}$ & $82.4_{1.6}$ & $82.3_{0.9}$ & $81.9_{2.1}$\\ 
MNLI-MM & $86.6_{0.8}$ & $82.6_{2.8}$ & $83.5_{1.4}$ & $82.6_{0.8}$ & $82.1_{2.0}$\\ 
\bottomrule
\end{tabular}
\caption{Mean and standard deviation (in subscript) of F-1 scores of models fine-tuned for NLI using the method proposed by \citet{pruthi2020weakly} with different base models (BERT and ELECTRA) and increasing proportion of examples with feature feedback. Results highlighted in bold are significant difference with $p<0.05$. \label{tab:nli1}}
\end{table*}

\section{Experiments}

We first fine-tune BERT and ELECTRA
on the annotated IMDb dataset \citep{zaidan2007using} 
following both classify-and-extract
and extract-then-classify approaches.
We evaluate resulting models on 
both iid test set as well as various 
naturally occurring out-of-domain datasets 
for sentiment analysis and compare resulting performance
with classify-only models (Table \ref{tab:sentiment1}).
We find that both approaches 
lead to significant gains 
(when tested with t-test with $p < 0.05$)
in out-of-domain performance 
compared to the classify-only method.
For instance, ELECTRA fine-tuned 
using the extract-then-classify framework 
leads to $\approx15.7\%$ gain in accuracy 
when evaluated on Yelp.
For NLI, however, training with rationales
doesn't lead to any visible performance gain (Table \ref{tab:nli2}).

As \citet{pruthi2020weakly} demonstrate
better performance on evidence extraction for sentiment analysis
compared to \citet{jain2020learning}, 
we use their method for additional analysis.
For both sentiment analysis and NLI, 
we fine-tune models with varying
proportion of samples with rationales
and report iid and out-of-domain performance
(Tables \ref{tab:sentiment2} and \ref{tab:nli1}).
% We compare models trained with rationales 
% against models trained without rationales
% and boldface the significant differences
% with p-value $<0.05$.
Training with no feature feedback 
recovers the classify-only baseline.

\begin{table}[t!]
\centering
\begin{tabular}{l c c}
\toprule
% & \multicolumn{3}{c}{Dataset size}\\
Test set & Classify-only & \citeauthor{zaidan2007using} \\
\midrule
% \multicolumn{5}{c}{SVM without/with rationales}\\
% \midrule
In-domain                                      
& $75.2_{3.5}$ & $79.1_{3.4}$ \\

CRD                                                               
& $48.3_{2.0}$ & $\mathbf{58.2_{2.4}}$ \\ 

SST-2
& $49.7_{0.3}$ & $\mathbf{65.6_{1.5}}$\\

Amazon
& $50.9_{0.3}$ & $\mathbf{68.7_{3.1}}$ \\ 

Semeval
& $49.8_{0.1}$ & $\mathbf{58.0_{1.5}}$ \\

Yelp
& $55.7_{2.8}$ & $\mathbf{74.8_{2.7}}$ \\
% \midrule
\bottomrule
\end{tabular}
\caption{ Mean and standard deviation (in subscript) of accuracy scores of classify-only SVM model versus SVM trained with feature feedback for sentiment analysis using \citet{zaidan2007using}'s method. Significant results ($p<0.05$) compared to the classify-only models are highlighted in bold.
}
\label{tab:sentiment3_only1200}
% \vspace{-5mm}
\end{table}

\begin{table}[t!]
\centering
\begin{tabular}{l c c}
\toprule
Task & Unigram & Bigram \\ 
\midrule
Sentiment Analysis & $21.37$ & $11.20$ \\
NLI & $79.54$ & $35.49$\\
\bottomrule
\end{tabular}
\caption{Percentage of unigram and bigram vocabularies that are marked as feature feedback at least once.}
\label{tab:vocab-prop}
\end{table}

\begin{table}[t!]
\centering
\begin{tabular}{l c c}
\toprule
 & Entailment & Contradiction \\ 
\midrule
$D\textsubscript{all}$ & $0.25$ & $0.16$ \\
$D\textsubscript{rationale}$ & $0.30$ & $0.09$\\
\bottomrule
\end{tabular}
\caption{Mean Jaccard index of premise-hypothesis word overlap ($D\textsubscript{all}$) and rationale overlap ($D\textsubscript{rationale}$) in the training set.}
\label{tab:jaccard}
\end{table}

On sentiment analysis, 
we find feature feedback to improve BERT's iid performance 
but find ELECTRA's performance comparable 
with and without feature feedback.
Feature feedback leads to an increase 
in performance out-of-domain on both BERT and ELECTRA.
For instance, with feature feedback,
% for all training examples, 
ELECTRA's classification accuracy 
increases from 
$91.6\%$ to $93.7\%$ on CRD and 
$79\%$ to $84.6\%$ on Yelp.
Similar trends are also observed 
when we fine-tune BERT with feature feedback.
Interestingly, when evaluated 
on the SemEval dataset (Tweets),
we observe that BERT 
fine-tuned with feature feedback 
on all training examples
achieves comparable performance to
fine-tuning without feature feedback.
However, fine-tuning with feature feedback 
on just $25\%$ of training examples 
leads to a significant improvement 
in classification accuracy.
We speculate that this might be a result of 
implicit hyperparameter tuning
when combining prediction and extraction losses,
and a more extensive hyperparameter search
could provide comparable (if not better) 
gains with $100\%$ data.
% For both BERT and ELECTRA, we also find 
% that fine-tuning with feature feedback 
% leads to a lower variance 
% (across different random initializations) 
% in model accuracy when evaluated out-of-domain.
Similarly, SVM trained with feature feedback \citep{zaidan2007using}
consistently outperformed SVM 
trained without feature feedback,
when evaluated out-of-domain 
despite obtaining similar accuracy
in-domain 
(Table~\ref{tab:sentiment3_only1200} and Appendix Table \ref{tab:sentiment3}).
For instance, SVM trained on just label information
% from $1200$ training examples 
achieved $75.2\% \pm 3.5\%$ accuracy 
on the in-domain test set, 
% and SVM trained with label information 
% along with feature feedback 
% achieved comparable accuracy of $79.1\% \pm 3.4\%$.
which was comparable to the accuracy of $79.1\% \pm 3.4\%$
achieved by SVM trained with feature feedback.
But the classifier trained with feature feedback 
led to $\approx 19\%$ and $\approx 18\%$ improvement 
in classification accuracy on Yelp reviews 
% and $\approx 18\%$ improvement on
and Amazon reviews, respectively,
compared to the classifier trained without feature feedback.

For NLI, it appears that 
feature feedback provides no added benefit 
compared to a classify-only BERT model,
whereas, ELECTRA's iid performance decreases with feature feedback.
Furthermore, models fine-tuned with feature feedback
generally perform no better
than classify-only models when trained with
varying proportions of rationales
(Table \ref{tab:nli1})
while classify-only models perform
significantly better than 
the models trained with rationales
when trained with varying dataset size.
(Appendix Table \ref{tab:nli2}).
These results are in line 
with observations in prior work on 
counterfactually augmented data \citep{huang2020counterfactually}.

\section{Discussion and Analysis}

\begin{table}[t!]
\centering
\begin{tabular}{l c c}
\toprule
Dataset &  \% Overlap & Label Agreement \\
\midrule
\multicolumn{3}{c}{Unigram}\\
\midrule
CRD & $60.3$ & $ 51.3$ \\
SST2 & $64.6$ & $66.5$ \\
Amazon & $45.6$ & $ 47.6$\\
Semeval & $30.9$ & $60.3$ \\
Yelp & $78.3$ & $65.1$\\ 
\midrule
\multicolumn{3}{c}{Bigram}\\
\midrule
CRD & $28.2$ & $51.9$ \\
SST2 & $28.5$ & $64.5$ \\
Amazon & $19.6$ & $49.9$\\
Semeval & $10.2$ & $58.5$ \\
Yelp & $46.8$ & $65.3$\\ 
\bottomrule
\end{tabular}
\caption{Rationale vocabulary overlap and label agreement between in-sample and OOD datasets. \label{tab:sa_vocab_overlap}}
\end{table}

% \begin{table}[t!]
% \centering
% \begin{tabular}{l c c}
% \toprule
% Task & Unigram & Bigram \\ 
% \midrule
% Sentiment Analysis & $21.37$ & $11.20$ \\
% NLI & $79.54$ & $35.49$\\
% \bottomrule
% \end{tabular}
% \caption{Percentage of unigram and bigram vocabularies that are marked as feature feedback at least once.}
% \label{tab:vocab-prop}
% \end{table}

% \begin{table}[t!]
% \centering
% \begin{tabular}{l c c}
% \toprule
%  & Entailment & Contradiction \\ 
% \midrule
% $D\textsubscript{all}$ & $0.25$ & $0.16$ \\
% $D\textsubscript{rationale}$ & $0.30$ & $0.09$\\
% \bottomrule
% \end{tabular}
% \caption{Mean Jaccard index of premise-hypothesis word overlap ($D\textsubscript{all}$) and rationale overlap ($D\textsubscript{rationale}$) in the training set.}
% \label{tab:jaccard}
% \end{table}

To further study the different trends 
on sentiment analysis versus NLI, 
we analyze feature feedback in both datasets.
We find that $21.37\%$ of tokens 
in the vocabulary of \citet{zaidan2007using} 
are marked as rationales 
in at least one movie review.
Interestingly, this fraction is
$79.54\%$ for NLI
% in the NLI training set
(Table \ref{tab:vocab-prop}).
% Movie reviews might contain certain words or phrases 
While for movie reviews, certain words or phrases might
generally denote positive or negative sentiment
(e.g., ``amazing movie''), 
for NLI tasks, 
% there is no notion of such words or phrases 
it is not clear that any individual phrase 
should suggest entailment or contradiction generally.
% Hence, it is likely that 
% Notably a word
A word or a phrase might be marked 
as indicating entailment in one NLI example
but as a contradiction in another. This
may explain why training with rationales
lead to no improvement in the NLI task.

We further construct vocabulary
of unigrams and bigrams from 
phrases marked as feature feedback 
in examples from the sentiment analysis training set 
($V\textsubscript{rationale}$).
We compute the fraction of unigrams (and bigrams)
that occur in this vocabulary
and also occur in each out-of-domain dataset.
We find that a large fraction
of unigrams from $V\textsubscript{rationale}$
also exist in CRD ($\approx60\%$), SST2 ($\approx64\%$),
and Yelp ($\approx 78\%$) data. 
(movie and restaurant reviews). 
However, this overlap is much smaller 
for SemEval ($\approx30\%$) 
and Amazon ($\approx45\%$)
, which consist of tweets and product reviews, 
respectively. 
For these overlapping unigrams, 
we observe a relatively large percentage 
($50$--$65\%$)
preserve their associated majority training set label 
in the out-of-domain datasets.
Similar trends hold for bigrams,
though fewer $V\textsubscript{rationale}$ bigrams
are present out-of-domain
(Table \ref{tab:sa_vocab_overlap}).
A model that pays more attention to these
spans might perform better out of domain.

For each pair in the NLI training set,
we compute Jaccard similarity 
between the premise and hypothesis sentence 
(Table \ref{tab:jaccard}).
We compute the mean of these example-level similarities 
over the entire dataset,
finding that it is common for examples in our training set
to have overlap between premise and hypothesis sentences,
regardless of the label.
However, when we compute mean Jaccard similarity 
between premise and hypothesis rationales, 
we find higher overlap
for entailment examples 
% labeled as entailment 
versus contradiction.
% (Jaccard index of $0.3$)
% and a significantly lower overlap 
% for contradictory pairs 
% (Jaccard index of $0.09$).
Thus, models trained with feature feedback
might learn to identify word overlap
as predictive of entailment 
even when the true label is contradiction.
While this may not improve an NLI model's performance,
it could be useful in tasks like Question Answering,
where answers often lie in sentences 
that have high word overlap with the question
\citep{lamm2020qed,majumder2021model}.
Interestingly, our results on NLI 
are in conflict with recent findings 
where models trained with rationales 
showed significant improvement
over classify-only models in both iid and out-of-domain
(MNLI-M and MNLI-MM) settings \citep{stacey2021natural}.
This
% explanation for these different findings 
could be due to the different modeling strategy 
employed in their work,
% While we guide models by extracting rationales, 
% \citet{stacey2021natural} 
% who do not 
% build an extractor model,
as they use rationales to 
guide the training of the classifier's attention module.
Investigating this difference is left for future work.

\section{Conclusion}
In this paper, we investigate the practical benefits of using feature feedback in two well-known tasks in NLP: sentiment analysis and natural language inference. Using two techniques that were primarily introduced for boosting interpretability as the basis of our experiments, we find they also have an unexpected advantage in boosting model robustness. Our experiments and analyses offer insight into how these interpretability methods may encourage generalization in out of domain settings.

To answer our first research question, we show that models trained with feature feedback can lead to performance improvement in the sentiment analysis task but not in NLI. 
To answer our second question, we find that as little as 25\% of the dataset can achieve the best performance in the out-of-domain setting in sentiment analysis, whereas no clear trends are visible in NLI.
Our analysis reveals that a smaller percentage of vocabulary is selected as rationales in sentiment analysis compared to NLI, indicating rationale tokens in the sentiment analysis task contain more distinctive information than NLI. Rationale tokens are more likely to exist among entailment samples than contradiction, which may lead the model to correlate the existence of rationales with entailment. 

% more reference to research questions we ask in intro (L99-114)
% future work
% again tie into why this is apt for bbnlp - 

\bibliography{anthology,emnlp2021}
\bibliographystyle{acl_natbib}

\newpage

\appendix

\section{Appendix}
\label{sec:appendix}

\begin{table*}[t!]
\centering
\small
  \begin{tabularx}{16cm}{lX}
    \toprule
    Task & Examples \\
    \midrule
     Sentiment Analysis (Positive) & $\dots$ characters are portrayed with such saddening realism that you can't help but love them , as pathetic as they really are . although levy \hl{stands out} , guest , willard , o'hara , and posey \hl{are all wonderful and definitely should be commended for their performances ! if there was an oscar for an ensemble performance , this is the group that should sweep it} $\dots$      \newline
     \\
     Sentiment Analysis (Negative) & $\dots$ then , as it's been threatening all along , the film explodes into violence . and just when you think it's finally over , schumacher \hl{tags on a ridiculous self-righteous finale that drags the whole unpleasant experience down even further} . trust me . \hl{there are better ways to waste two hours of your life} $\dots$
     \newline
    \\
    NLI (Entailment) &  \textbf{P:} a \hl{white dog drinks water} on a mountainside.
    \newline \textbf{H:} there is a \hl{dog drinking water} right now.
         \newline
    \\
    NLI (Contradiction) &  \textbf{P:} a dog \hl{leaping off a boat}
    \newline \textbf{H:} dogs \hl{drinking water} from pond 
    \\
\bottomrule
  \end{tabularx}
  \caption{Examples of documents (and true label) with feature feedback (highlighted in yellow).}
  \label{tab:train-examples}
\end{table*}

\begin{table*}[t!]
\centering
\small
  \begin{tabularx}{16cm}{lX}
    \toprule
    Task & Examples \\
    \midrule
     Sentiment Analysis (Positive, Correct) & everyone should adapt a tom robbins book for screen . while the \hl{movie is fine} and \hl{the performances are good} , the dialogue , which \hl{works well reading} it , \hl{is beautiful} when spoken .
     \newline\\
     Sentiment Analysis (Positive, Wrong) &  ... \hl{very uncaptivating} yet one gets the feeling that their is some serious exploitation going on here ... 
     \newline \\
     Sentiment Analysis (Negative, Correct) &  ... using quicken is \hl{a frustrating experience} each time i fire it up ... 
     \newline\\
     Sentiment Analysis (Negative, Wrong) &  ... with many \hl{cringe-worthy} `surprises', which happen around 10 minutes after you see exactly what's going to happen ...
     \newline\\
    NLI (Entailment, Correct) &  \textbf{P:}  a woman cook in an apron is \hl{smiling at the camera} with two other cooks in the background .
    \newline  \textbf{H:} a woman \hl{looking at the camera .}
    \newline \\
    NLI (Entailment, Wrong) &  \textbf{P:} a \hl{woman} in a \hl{brown dress} looking at papers in front of a class .
    \newline  \textbf{H:} a woman looking at papers in front of a class is \hl{not wearing a blue dress .}
    \newline \\
    NLI (Contradiction, Correct) &  \textbf{P:} the woman \hl{in} the \hl{white dress} looks very uncomfortable in the busy surroundings 
    \newline  \textbf{H:} the \hl{dress is black .}
    \newline \\
    NLI (Contradiction, Wrong) &  \textbf{P:} a \hl{man} , wearing a cap , is \hl{pushing a cart , on} which \hl{large display boards} are kept , on a road .
    \newline  \textbf{H:} the \hl{person} is \hl{pulling large display boards on} a \hl{cart .} 
    \newline \\
\bottomrule
  \end{tabularx}
  \caption{Examples (from out-of-domain evaluation sets; with true label and model prediction) of explanations highlighted by feature feedback models (highlighted in yellow).}
  \label{tab:ood-examples}
\end{table*}

% \begin{table*}
% %   \vspace{-10cm}
% \centering
% \begin{tabular}{l c c}
% \toprule
% Task &  BERT & ELECTRA \\
% \midrule
% Sentiment Analysis & $45.8_{2.8}$ & $51.7_{0.7}$ \\
% NLI & $56.5_{0.4}$ & $59.2_{0.9}$ \\

% \bottomrule
% \end{tabular}
% \caption{Rationale extraction F1 scores of BERT and ELECTRA models
% trained with $100\%$ rationales. NLI 
% results are for models trained with 6318 training samples.\label{tab:extraction_f1}}
% %   \vspace{-20cm}
% \end{table*}

\begin{table*}[t!]
\centering
\begin{tabular}{l c c c c}
\toprule
& \multicolumn{3}{c}{Dataset size}\\
Evaluation Set &  $300$ & $600$ & $900$ & $1200$ \\
\midrule
% \multicolumn{5}{c}{SVM without/with rationales}\\
% \midrule
In-domain                                      
& $77.0_{3.9}$/$77.6_{2.2}$
& $78.5_{3.2}$/$82.3_{2.0}$
& $80.5_{1.7}$/$\mathbf{84.9_{1.6}}$
& $75.2_{3.5}$/$79.1_{3.4}$ \\

CRD                                                               
& $48.0_{2.9}$/$\mathbf{56.4_{1.3}}$
& $48.3_{2.5}$/$\mathbf{58.0_{2.7}}$
& $48.4_{2.3}$/$\mathbf{58.7_{1.8}}$
& $48.3_{2.0}$/$\mathbf{58.2_{2.4}}$ \\ 

SST-2                                                                                  
& $52.2_{1.6}$/$\mathbf{62.9_{1.0}}$
& $50.9_{3.0}$/$\mathbf{64.0_{0.9}}$
& $51.3_{3.1}$/$\mathbf{64.9_{0.9}}$
& $49.7_{0.3}$/$\mathbf{65.6_{1.5}}$\\

Amazon                                             
& $51.8_{1.5}$/$\mathbf{65.9_{1.9}}$
& $52.4_{2.0}$/$\mathbf{66.5_{1.2}}$
& $52.0_{2.9}$/$\mathbf{69.9_{0.4}}$
& $50.9_{0.3}$/$\mathbf{68.7_{3.1}}$ \\ 

Semeval                                                                                
& $50.3_{1.4}$/$\mathbf{56.7_{1.1}}$
& $50.3_{1.2}$/$\mathbf{56.4_{0.8}}$
& $50.1_{0.5}$/$\mathbf{58.8_{1.3}}$
& $49.8_{0.1}$/$\mathbf{58.0_{1.5}}$ \\

Yelp                                                                                
& $60.2_{4.0}$/$\mathbf{72.0_{2.4}}$
& $57.3_{7.1}$/$\mathbf{74.5_{1.5}}$
& $61.2_{4.6}$/$\mathbf{74.8_{2.5}}$
& $55.7_{2.8}$/$\mathbf{74.8_{2.7}}$ \\
% \midrule
\bottomrule
\end{tabular}
\caption{ Mean and standard deviation (in subscript) of accuracy scores of classify-only SVM models (left) presented alongside accuracy scores of models trained with feature feedback (right), with increasing number of training-samples for sentiment analysis using the method proposed by \citet{zaidan2007using}. Results highlighted in bold show statistically significant difference with $p<0.05$.
}
\label{tab:sentiment3}
\end{table*}

\begin{table*}[t!]
\centering
\begin{tabular}{l c c c c}
\toprule
& \multicolumn{3}{c}{Dataset size}\\
Evaluation Set &  $1500$ & $3000$ & $4500$ & $6318$ \\
\midrule
\multicolumn{5}{c}{BERT}\\
\midrule
In-domain                                      
& $85.9_{6.0}$/$84.5_{2.0}$ 
& $87.9_{0.4}$/$87.7_{1.0}$  
& $89.1_{0.4}$/$89.2_{0.2}$
& $88.7_{2.0}$/$89.8_{0.8}$ \\

RP                                                               
& $61.8_{0.9}$/$62.8_{1.8}$ 
& $63.3_{1.6}$/$64.2_{1.8}$  
& $63.7_{1.8}$/$\mathbf{66.8_{1.4}}$  
& $62.9_{3.9}$/$66.4_{1.7}$ \\ 
RH                                                                                  
& $74.5_{1.6}$/$71.8_{3.4}$ 
& $77.0_{1.4}$/$77.3_{2.1}$  
& $78.3_{1.1}$/$80.4_{1.8}$ 
& $76.9_{3.5}$/$80.5_{1.9}$\\

MNLI-M                                              
& $63.7_{3.1}$/$60.8_{3.2}$ 
& $69.2_{1.8}$/$66.3_{2.2}$ 
& $70.2_{0.9}$/$67.5_{3.1}$  
& $69.7_{2.6}$/$68.1_{1.9}$ \\ 

MNLI-MM                                                                                    
& $64.8_{4.3}$/$61.8_{4.3}$  
& $71.3_{2.3}$/$67.5_{2.8}$ 
& $72.1_{1.2}$/$68.9_{4.2}$   
& $73.1_{1.9}$/$71.4_{1.1}$ \\
\midrule
\multicolumn{5}{c}{ELECTRA}\\
\midrule
In-domain 
&$\mathbf{94.6_{0.2}}$/$92.7_{0.5}$ 
& $\mathbf{95.1_{0.4}}$/$94.2_{0.3}$
& $\mathbf{95.7_{0.2}}$/$94.4_{0.2}$
& $\mathbf{96.0_{0.2}}$/$95.1_{0.3}$ \\

RP
& $78.4_{1.2}$/$75.2_{2.5}$ 
& $78.5_{1.8}$/$77.2_{0.9}$  
& $\mathbf{81.2_{0.6}}$/$76.2_{1.2}$   
& $\mathbf{80.8_{1.0}}$/$78.0_{0.6}$ \\ 

RH    
& $\mathbf{87.7_{0.7}}$/$85.2_{1.4}$ 
& $88.1_{1.3}$/$87.3_{0.6}$  
& $\mathbf{89.4_{0.6}}$/$87.1_{1.0}$ 
& $88.9_{1.0}$/$88.7_{0.9}$\\

MNLI-M  
& $\mathbf{82.8_{2.2}}$/$77.0_{1.8}$ 
& $\mathbf{85.4_{1.8}}$/$78.9_{1.7}$ 
& $\mathbf{86.0_{1.6}}$/$80.4_{2.1}$  
& $\mathbf{86.5_{0.9}}$/$81.9_{2.1}$ \\ 

MNLI-MM                                                                 
& $\mathbf{83.6_{2.5}}$/$77.9_{2.1}$  
& $\mathbf{86.2_{2.1}}$/$79.9_{1.9}$ 
& $\mathbf{86.1_{1.8}}$/$80.8_{2.2}$   
& $\mathbf{86.6_{0.8}}$/$82.1_{2.0}$ \\ 
\bottomrule
\end{tabular}
\caption{ Mean and standard deviation (in subscript) of F-1 scores of classify-only models/models trained with feature feedback, with increasing number of training-samples for NLI using the method proposed by \citet{pruthi2020weakly}. Results highlighted in bold are statistically significant difference with $p<0.05$.} 
\label{tab:nli3}
\end{table*}

% This is an appendix.

\end{document}